\documentclass[a4paper,10pt]{amsart}
\usepackage[utf8]{inputenc}

\usepackage{amsmath,amsthm,amssymb}
\usepackage{latexsym}
\usepackage{geometry}
\usepackage{xcolor}
\usepackage{float} 
\usepackage{bm}
\usepackage{todonotes}

\title[All's well that FID's well?]{All's well that FID's well? \\ Result quality and metric scores in GAN models for lip-sychronization tasks}

\author{Carina Geldhauser}
\address{Lund University, Centre for Mathematical Sciences, Box 118, 22100 Lund, Sweden.}
\email{carina.geldhauser@math.lth.se}
\author{Johan Liljegren}
\email{tna15jli@student.lu.se}
\author{Pontus Nordqvist}
\email{tna15pno@student.lu.se}

\begin{document}

\begin{abstract}
We test the performance of GAN models for lip-synchronization. For this, we reimplement LipGAN in Pytorch, train it on the dataset GRID and compare it to our own variation, L1WGAN-GP, adapted to the LipGAN architecture and also trained on GRID. 
\end{abstract}

\maketitle



\section{Introduction}
\label{sec:intro}

Facial animation is an important element in 
Computer Generated Imagery. Humans read off a variety of information about the character and the scene from facial expressions, from emotions to situations of joy, tension or danger. 
Also, human perception is very sensitive to anomalies in facial motion or dissynchronization between audio and visual information.
This makes the task of generating realistic videoclips, even just with a talking face, very challenging: It requires high-quality faces, lip movements synchronized with the audio, and plausible facial expressions.  
Traditional approaches to facial synthesis in Computer Generated Imagery can already produce faces that exhibit a high level of realism. However, most traditional approaches such as mouth shapes \cite{simons1990generation} or 3D meshes are speaker-specific and therefore need to be re-arranged for every new character, requiring a huge amount of video footage of the target person for training, modeling, or sampling. Due to the  need for expensive equipment and high amount of specialist work, such projects are still mostly undertaken by large studios. 

\begin{figure}
    \centering
    \includegraphics[scale = 0.52]{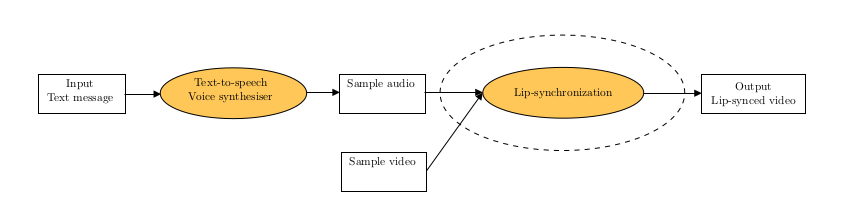}
   \caption{A sketch of our vision of a pipeline for a personalized video message service. This work focuses on lip-synchronization which is marked by the dashed circle.}
   \label{fig:pipe}%
\end{figure}

In order to drive down the cost and time required to produce high quality computer generated audio-visual sceneries,
researchers are looking into automatic face synthesis using machine learning
techniques. 
Lip-sychronization is a task of particular interest, 
since speech acoustics are highly correlated with facial movements.
Applications include film animation processes or  movie dubbing, and other steps of post-production with the scope to achieve better lip-synchronization. 
Finally, this technology can improve band-limited visual telecommunications
by either generating the entire visual content based on the audio or filling in dropped frames, which was precisely our motivation for this work.  The pipeline for visual content creation in band-limited telecommunications is as displayed in figure \ref{fig:pipe}.

In this exploratory work, we investigate a machine learning model used for transfering speech information to an arbitrary photo in the wild: given a single face image, use a suitably trained GAN to generate a short video clip of this face talking,  based on speech information given through an audio file. We exemplify this on two models, LipGAN and WGAN-GP, and evaluate some common metrics for visual quality and their (non-)agreeance with human eye inspection.

\section{Related Work}

The task of synthesizing lip motion is classical in computer vision and computer graphics, where a plethora of works deal with synthesizing lip motion from either audio  \cite{xie2007realistic,wang2010synthesizing}
or generating talking faces from videos \cite{thies2016face2face}.
A famous recent example is \cite{suwajanakorn2017synthesizing}, who exemplified the power of their traditional computer vision technique on photo-realistic generated videoclips of the former president of the United States, Barack Obama.   The author emphasizes the problem of creating a credible fake video due to the human attentiveness to details in the mouth area. 
Its extension ObamaNet \cite{kumar2017obamanet}, which integrates text-to-voice synthesizing to the model, then uses a recurrent neural networks for audio processing and U-Nets for video processing. 

In recent years, models based on neural networks 
\cite{fan2015photo,liu2017video,wiles2018x2face} became more popular, in particular due to their potential to generate arbitrary-identity talking faces.
Recent GAN-based approaches to produce lip-synchronizing methods include \cite{yao2021iterative}, who use a phoneme search approach to target mumbling and unwanted words in videos to remove them; the supervised learning approach \cite{zhou2019talking} uses labeled audio and video from different persons to create audio- and video embeddings, which are combined by a temporal GAN. 
Several implementations of a data-driven unsupervised learning approach have been made, including \cite{chung2017said} that uses encoders for audio and video separately which are then concatenated to a single embedding. This embedding is then decoded by a single decoder to produce a video. Similarly, \cite{chen2018} uses encoder and decoders in the same manner but with a more advanced pipeline that uses more networks. 
In this work, we examine \cite{LipGANArticle}, a model aiming to perform automated translation of videos, utilizing a GAN in combination with autoencoders, and compare it to our own implementation based on Wasserstein GANs.

\section{Datsets}

\begin{figure}
  \centering
  \begin{tabular}{@{}c@{}}
    \includegraphics[width=0.45\linewidth]{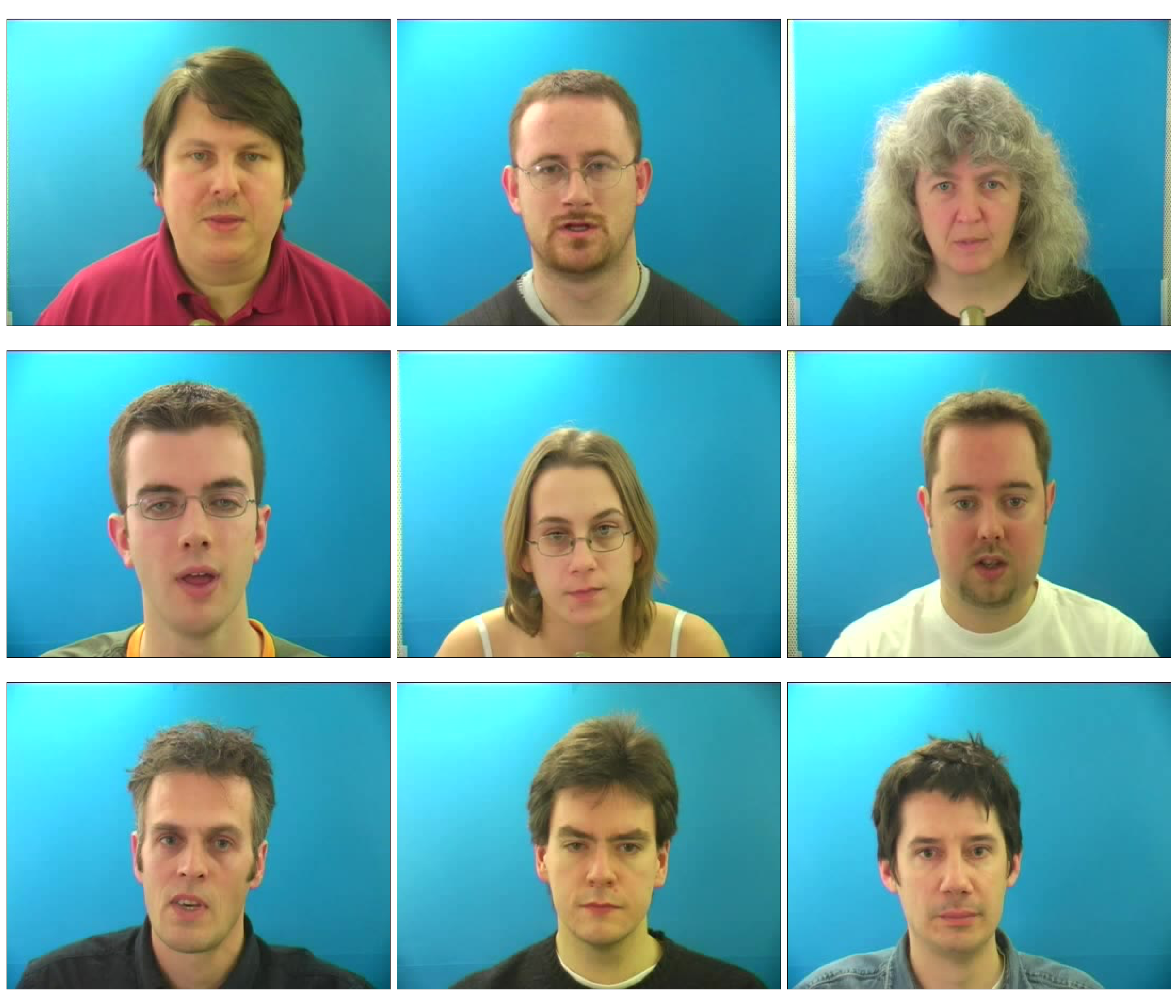} \\[\abovecaptionskip]
    \small (a) GRID dataset
  \end{tabular}
  \begin{tabular}{@{}c@{}}
    \includegraphics[width=.45\linewidth]{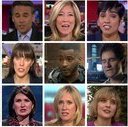} \\[\abovecaptionskip]
    \small (b) LRS2 dataset
  \end{tabular}
      \caption{Samples of speakers from the datasets GRID \cite{cooke2006audio} and LRS2 \cite{Afouras_2019}. Note that the resolution in the LRS2 samples is not representative since it is a screenshot.}
 \label{fig:datasetsample}
\end{figure}

In this section we give a brief description of our main dataset GRID and compare it to LRS2, the dataset used in   \cite{LipGANArticle}.
Nine sample images from the respective datasets can be seen in figure \ref{fig:datasetsample}.

The  GRID dataset was introduced by \cite{cooke2006audio} as a corpus for tasks such as speech perception and speech recognition. GRID contains 33 unique speakers, articulating 1000 word sequences in separate videos, each about 3 seconds long. The total length of GRID video material is about $27.5$ hours of 25 frames per second video with a resolution of $360 \times 288$ and a bitrate $1$ kbit/s. Out of the 33 speakers, 16 were female and 17 were male, and all speakers had English as their first language. The videos are filmed in a lab environment with a green screen background rendering a clinic setting for the videos. The speakers are always faced forward and looking into the camera.

The authors of \cite{LipGANArticle} used about $29$ hours of the  LRS2 dataset \cite{Afouras_2019} to train their model. LRS2 consists of news recordings from the BBC, with different lighting, backgrounds, face poses, and people with different origins, making the LRS2 being more of a \textit{in-the-wild} dataset that captures real conversations.

 \noindent \textbf{Preprocessing of the data.} We cut each video into frames, crop out the face of each image and rescale it; separately, we divide the audio into audio segments in mel-spectrogram representation. 
To make our experiments comparable, we used the preprocess and audio code
from \cite{LipGANArticle}. 
All pre-processing resulted in training data which consisted of the real face inputs $S$, and the shifted frame $S'$, which had been resized to size $96 \times 96\times 3$ i.e $H=96$.
Furthermore, the shifted frames $S'$ were obtained by picking a frame using a time step $\pm \alpha$, where $\alpha$ is of random size $\alpha = 1,2,...,6$.
As for the audio data, it consisted of mel-spectrograms with $M=80$ mel-frequency channels, and a time window of $T=27$. This time window equivalates to about 300 ms of total audio, which is spread out evenly before and after the frame. 
Resulting data attributes are summarized in table \ref{tab:BaselineDataAttributes}.
\begin{table}[H]
\centering
\caption{Data attributes for the training data.}
\begin{tabular}{cc}
\hline
\hline
\multicolumn{2}{c}{\textbf{Data attributes}}                                   \\ \hline
\multicolumn{1}{c|}{Input image horizontal/vertical dimension $H$} & 96        \\
\multicolumn{1}{c|}{Frameshift time step $\alpha$}                 & $1,2,...,6$         \\
\multicolumn{1}{c|}{Mel-frequency channels $M$}                    & 80        \\
\multicolumn{1}{c|}{Mel-spectrogram time window $T$}               & 27        \\ \hline \hline
\end{tabular}
\label{tab:BaselineDataAttributes}
\end{table}
\noindent
Finally, all the pre-processed data resulted in 2202106 frames of faces, together with 33000 mel-spectrograms. This was then subdivided into the three sub-datasets GRIDSmall, GRIDFull, and GRIDTest. The first two subsets, GRIDSmall and GRIDFull, contained 300 and 980 video samples respectively from each of the 33 speakers. As the name suggests, the latter subset, GRIDTest, was used to test the models and therefore had no intersection of data with the two previously mentioned sub-datasets, which are used for training. Further, the test datasets contained 43929 image samples, which was specifically chosen since it matches the sample sizes used to calculate some specific GAN metrics, similar to other GAN comparison articles \cite{kurach2019largescale,chong2020effectively}. All used sub-datasets are summarized in table \ref{tab:DataSubsets}.
\begin{table}[H]
\centering
\caption{Information about the data subsets used for all experiments.}
\begin{tabular}{c|ccc}
\hline
\hline
\textbf{Name} & \textbf{Type} & \textbf{Individual samples} & \textbf{Videos per speaker} \\ \hline
GRIDSmall       & Train         &          670758                   & 300                         \\
GRIDFull      & Train         &           2190517                  & 980                         \\
GRIDTest      & Test          &        44589                     & 20                          \\ \hline \hline
\end{tabular}
\label{tab:DataSubsets}
\end{table}

\section{Experiment overview}

In a series of experiments, we compare two models performing the task of lip-synchronization with the GRID dataset as training data:   LipGAN \cite{LipGANArticle}, re-implemented by us on Pytorch, and an adapted Wasserstein GAN with gradient penalty \cite{gulrajani2017improved}, abbreviated as L1WGAN-GP. Our choice to use a Wasserstein GAN was motivated by the empirical study \cite{xu2018evaluation}, which concluded that WGAN-GP performs the best under most of the metrics they used to test it. 

To this aim, we first analyze our implementations of LipGAN and L1WGAN-GP  for convergence and inspect sample images produced during training to ensure that the generated samples were convincing faces of satisfying perceptual quality, then we apply three different quantitative metrics and finally a qualitative assessment. 
Both models were trained for 20 epochs with a batch size of 128,  using the same initial random seed of \texttt{numpy.random.seed(10)}. No batch normalization was used in the discriminator, as it can be a problem for a WGAN, see section \ref{ss:EnforceLipschitz}. Similarly, the number of trainable parameters stayed the same, at 47424915 where 37087763 are for the generator and 10337152 for the discriminator. 
ADAM was used as optimizer for both networks, using an initial learning rate of $\eta=10^{-4}$ and the decay parameters $\beta_1=0.5$ and $\beta_2=0.9$.

\subsection{LipGAN}

In the following, with ``LipGAN'' we refer to our reimplementation of \cite{LipGANArticle} in Pytorch instead of the outdated Keras version used in the original implementation.
 This model builds a pipeline that inputs a video in the source language and translates it to a target language with correctly lip-synced lips for the target language. 
LipGAN inputs frames and audio from and input distribution $\mathbb{P}_z$ and outputs it as a generated lip-synced frame in the output distribution $\mathbb{P}_g$.

\begin{figure}
  \centering
  \begin{tabular}{@{}c@{}}
    \includegraphics[width=0.45\linewidth]{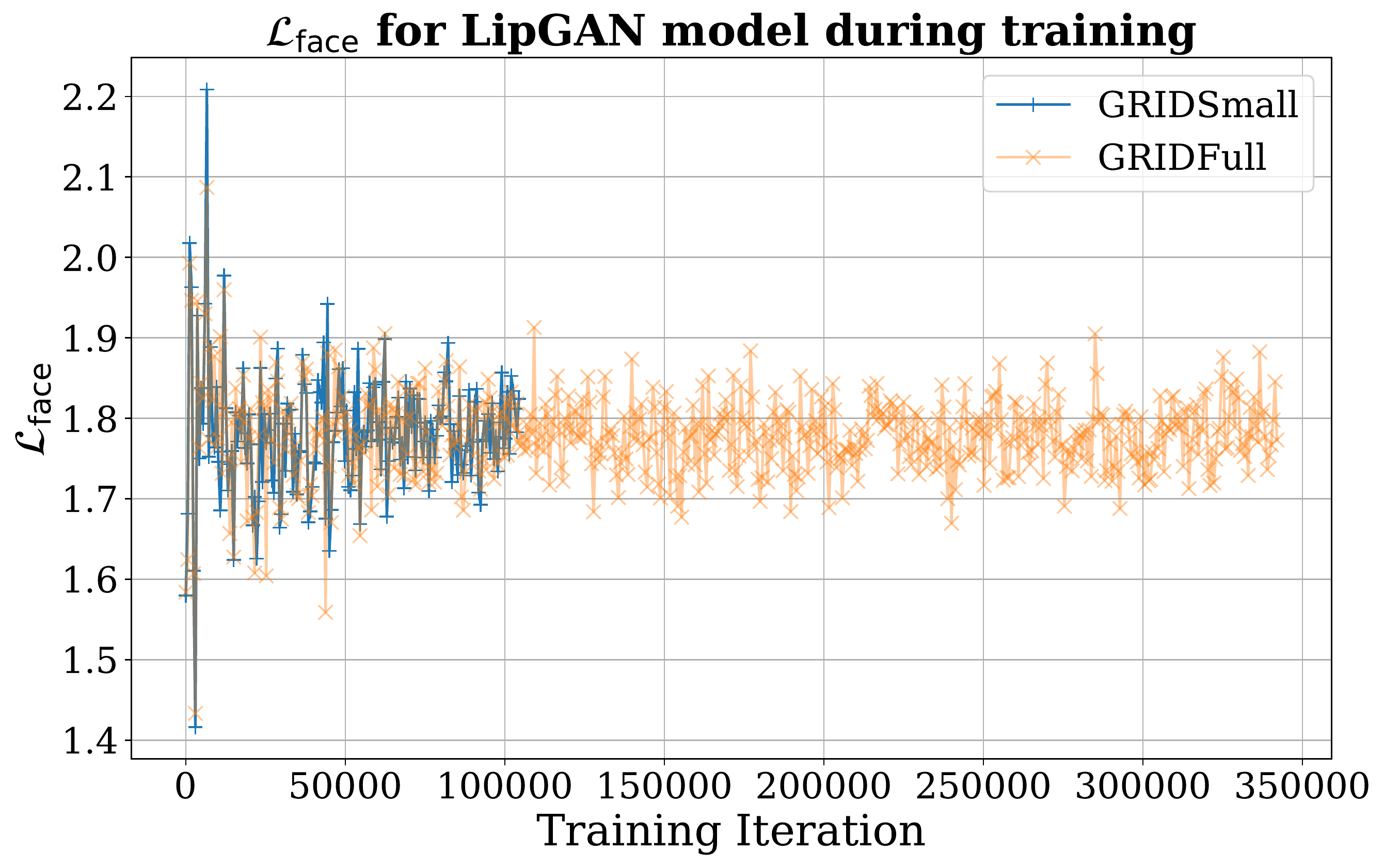} \\[\abovecaptionskip]
    \small (a) Faked face $\hat{S}$, synced audio $A$, loss $\mathcal{L}_{\mathrm{face}}$.
  \end{tabular}
  \begin{tabular}{@{}c@{}}
    \includegraphics[width=.45\linewidth]{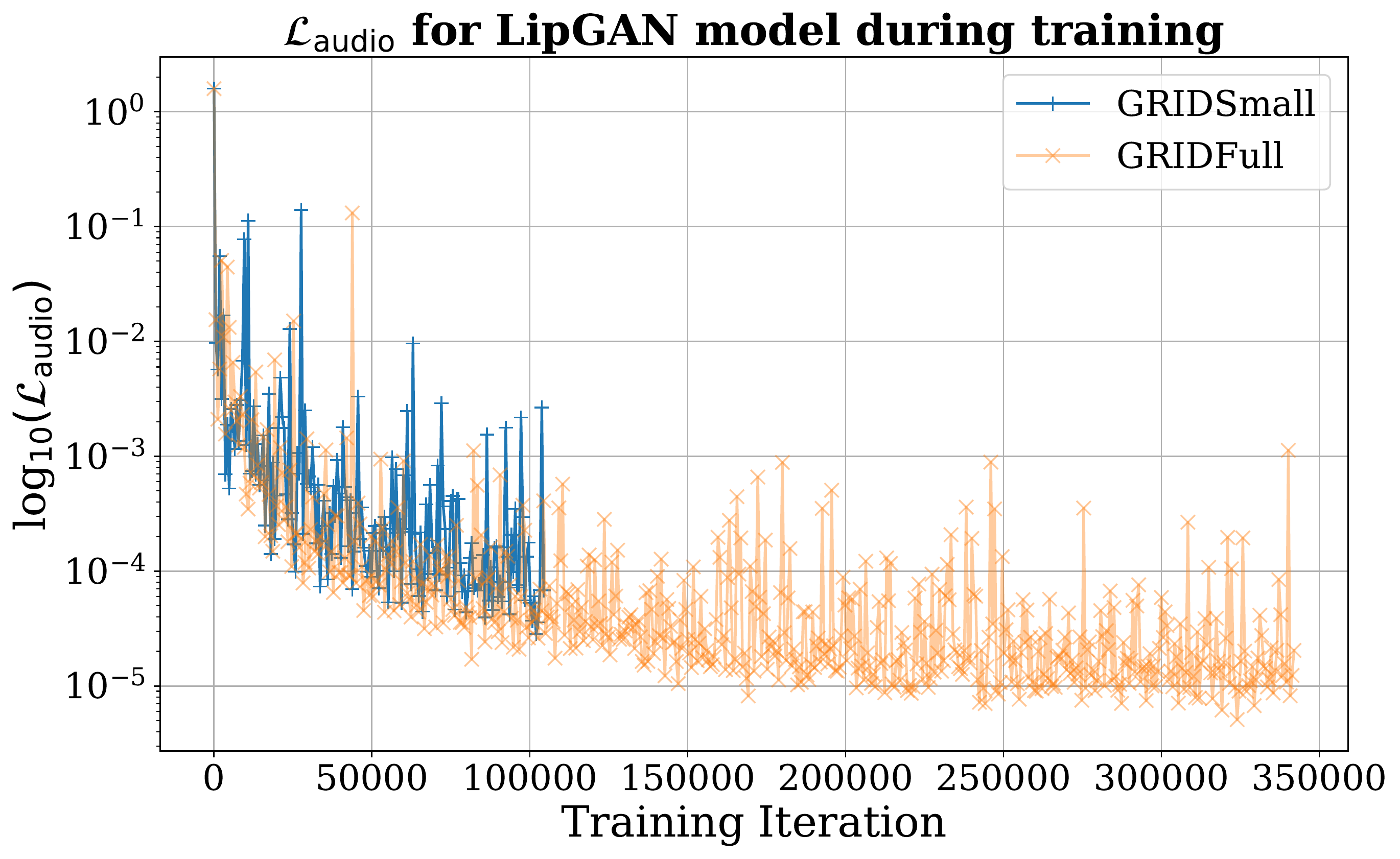} \\[\abovecaptionskip]
    \small (b) Real face $\hat{S}$, unsynced audio $A'$, loss $\mathcal{L}_{\mathrm{audio}}$.
  \end{tabular}
    \caption{Losses during training for the LipGAN model.}
    \label{fig:LipGANAuFaLoss}
\end{figure}

The LipGAN model was trained using the GRIDSmall and the GRIDFull datasets for 20 epochs. This took approximately 1 day with 105000 training iterations for GRIDSmall and 3 days with 342400 training iterations for GRIDFull, on both systems.
During training, the different LipGAN losses were sampled every 600th training iteration for both datasets.
\begin{figure}
  \centering
  \begin{tabular}{@{}c@{}}
    \includegraphics[width=0.45\linewidth]{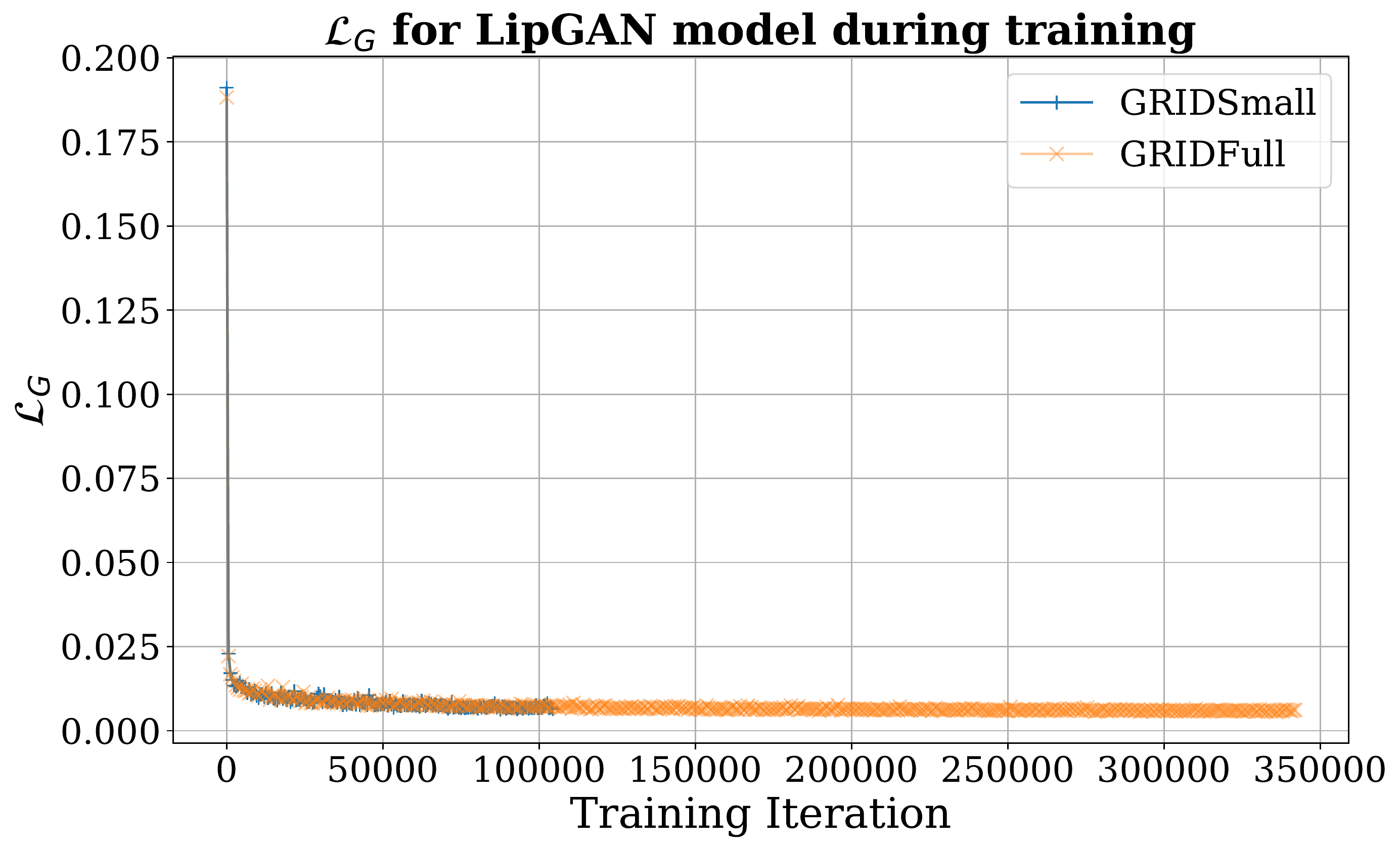} \\[\abovecaptionskip]
    \small (a) Generator loss $\mathcal{L}_G$
  \end{tabular}
  \begin{tabular}{@{}c@{}}
    \includegraphics[width=.45\linewidth]{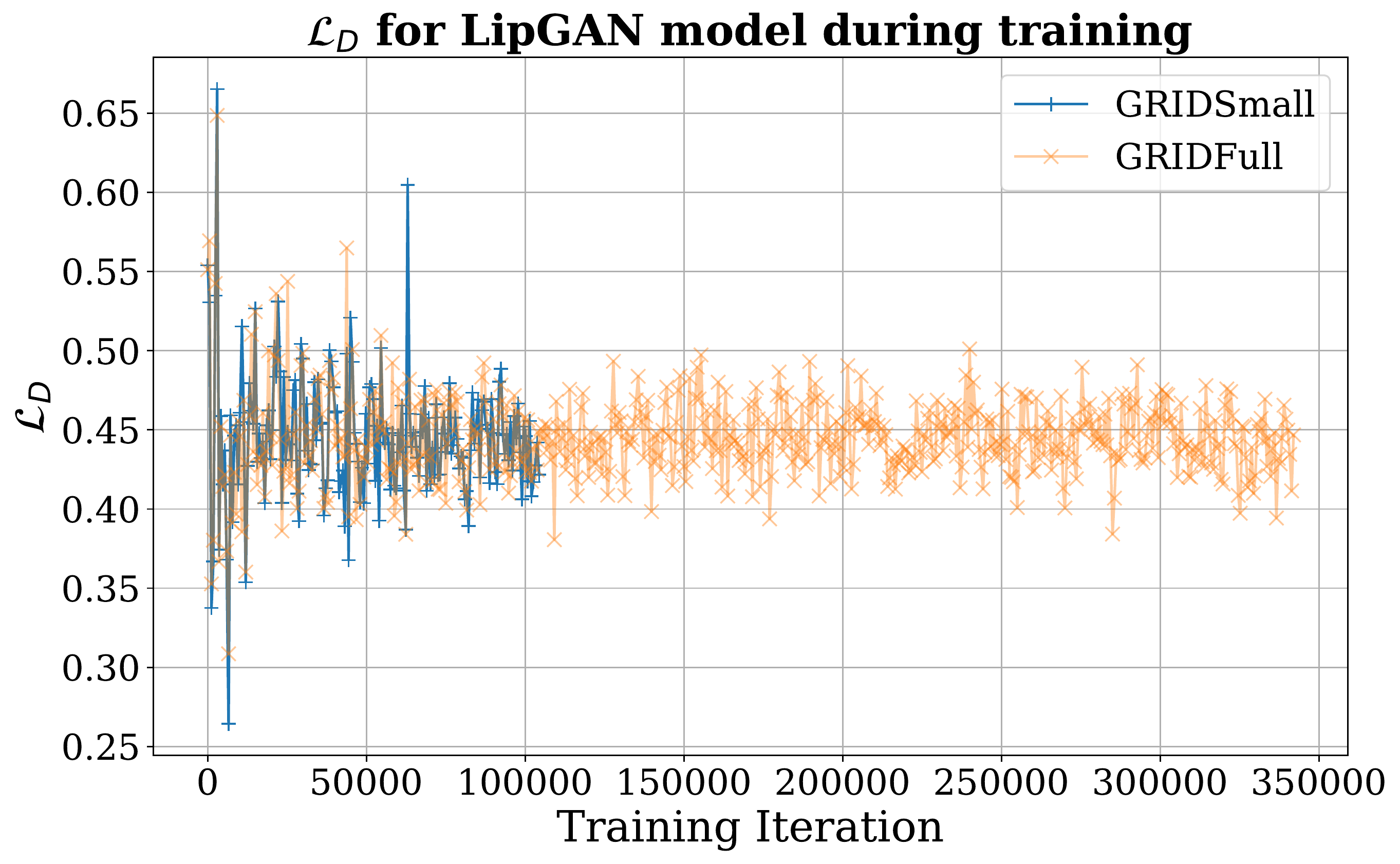} \\[\abovecaptionskip]
    \small (b) Critic loss $\mathcal{L}_D$
  \end{tabular}
\caption{Losses during training for the LipGAN model.}
    \label{fig:LipGANGenCriLoss}
\end{figure}
\noindent
As displayed in figure \ref{fig:LipGANGenCriLoss},
the generator loss $\mathcal{L}_G$ converges to around 0, with a minimum loss of $6.0\cdot 10^{-3}$ for GRIDSmall and $5.7\cdot 10^{-3}$ for GRIDFull. The discriminator loss $\mathcal{L}_D$  converges around 0.45, however, some outliers can be seen, which resulted in a search for potential errors in the training samples, although, none were found. Additionally, the loss $\mathcal{L}_{\mathrm{face}}$ with the input of a fake face $\hat{S}$ with real audio $A$, and the loss $\mathcal{L}_{\mathrm{audio}}$ with the input of a real face $S$ but time-unsynced audio $A'$, were sampled. This was done even if they did not contribute to the discriminator loss $\mathcal{L}_D$ for that specific iteration. 
These losses can be seen in figure \ref{fig:LipGANAuFaLoss}. 
Despite some outliers for the loss $\mathcal{L}_{\mathrm{audio}}$, both losses seem to converge, although, this process is slower for $\mathcal{L}_{\mathrm{audio}}$ than $\mathcal{L}_{\mathrm{face}}$.

\noindent

\begin{figure}
  \centering
  \begin{tabular}{@{}c@{}}
    \includegraphics[width=0.45\linewidth]{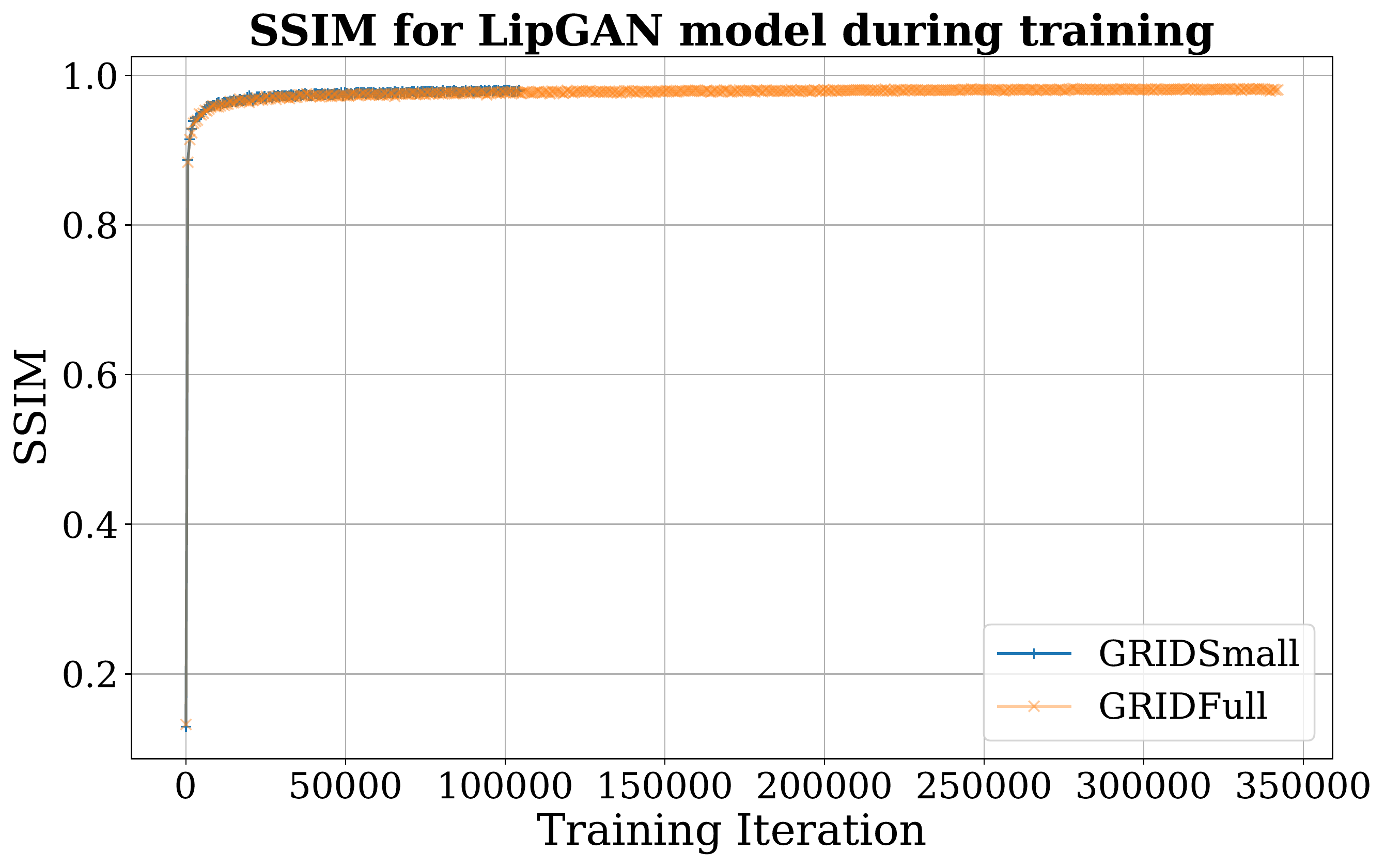} \\[\abovecaptionskip]
    \small (a) SSIM
  \end{tabular}
  \begin{tabular}{@{}c@{}}
    \includegraphics[width=.45\linewidth]{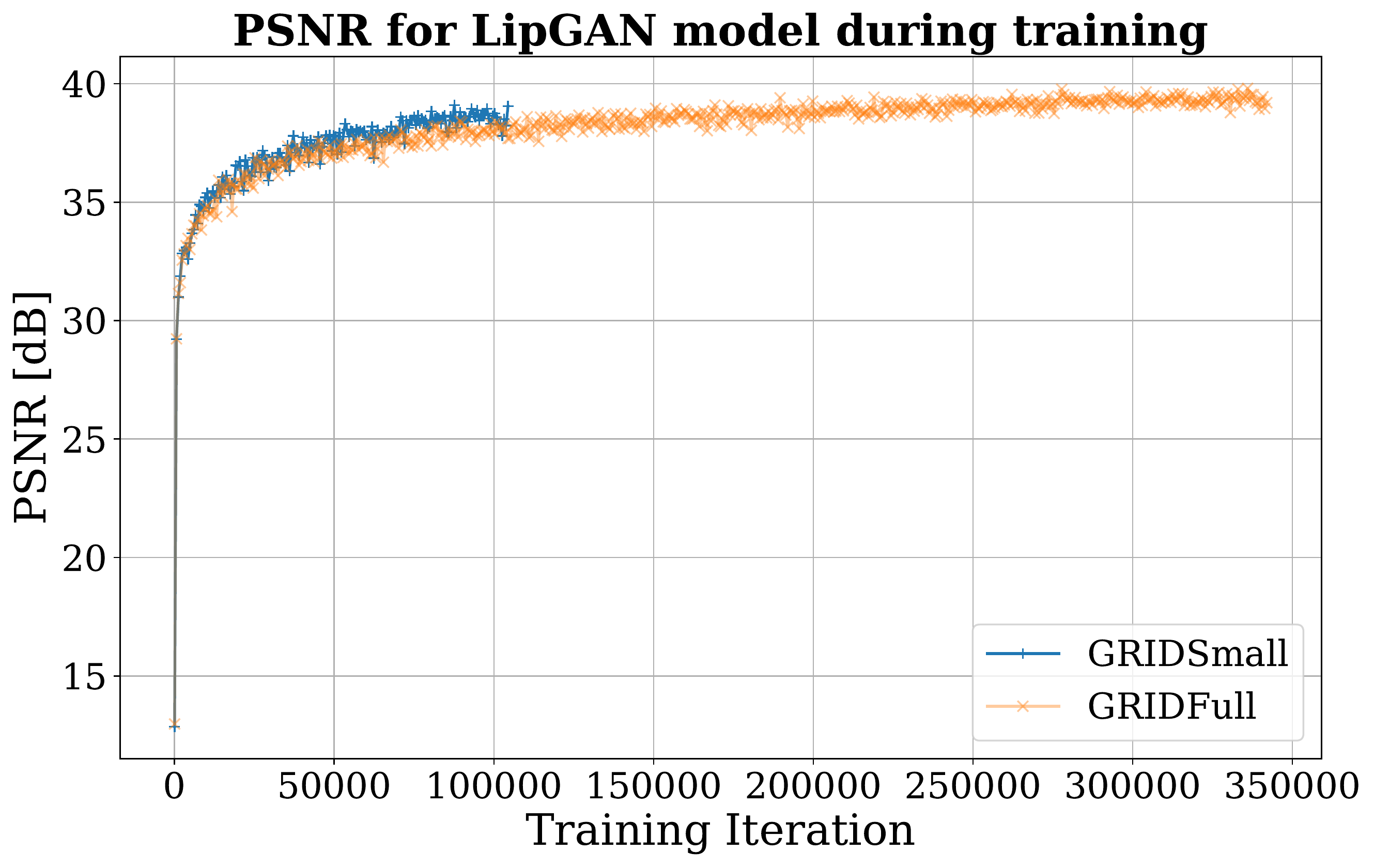} \\[\abovecaptionskip]
    \small (b) PSNR
  \end{tabular}
 \caption{The metrics SSIM and PSNR for the LipGAN model during training. The metrics have been taken for every 600th training iteration.}
    \label{fig:LipGANTrainingMetrics}
\end{figure}

\noindent \textbf{Sample inspection.} Most generated faces are of good quality after the first epoch. However, if one looks closely, small differences for some select samples can be seen. For example, in the sample for epoch 7, one can see that the generator produces a face with an open mouth, while the mouth is more closed for the ground truth face. Further, for sample 9, one can notice that the beard has a more blurry appearance than its ground truth counterpart. Lastly, SSIM and PSNR were calculated for the generator's samples, together with the ground truth counterpart, every 600th training iteration. This can be seen in figure \ref{fig:LipGANTrainingMetrics}.
\begin{figure}
    \centering
    \includegraphics[scale = 0.17]{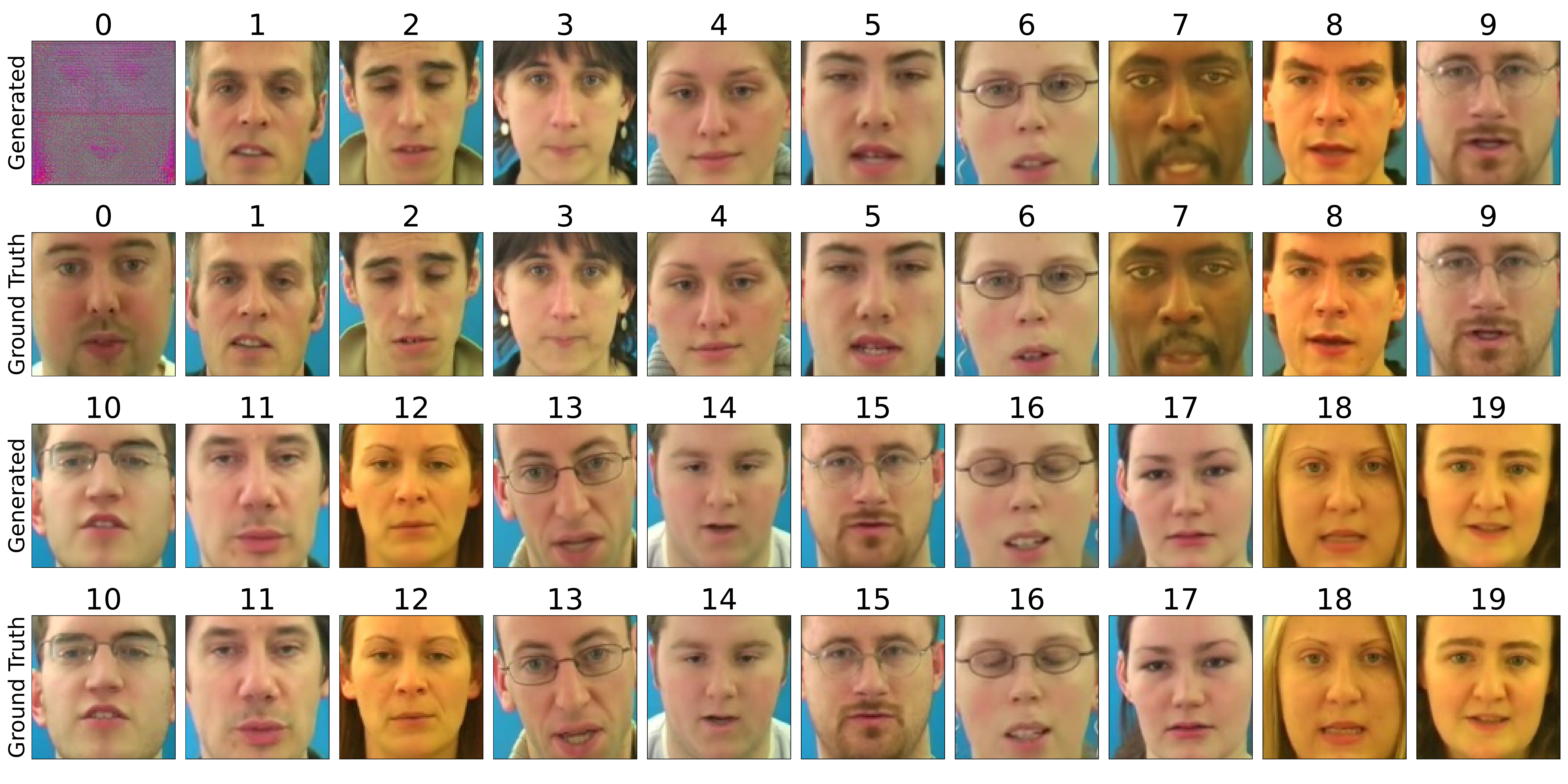}
    \caption{20 random generated faces $\hat{S}$ from the generator, together with their corresponding true faces $S$, from the training of the LipGAN model using GRIDFull. The number denotes the epoch.}
    \label{fig:LipGANSamples}
\end{figure}

\noindent \textbf{Quality metrics.}
For convenience, we plot  SSIM and PSNR scores for our LipGAN implementation, once with the reduced dataset GRIDSmall and once for GRIDFull.  SSIM ranges from around 0.13 to 0.98 for both datasets,  PSNR  starts at around 13 dB for both datasets, and end at about 39 dB for GRIDSmall and 40 dB for GRIDFull.
  
\subsection{L1WGAN-GP}\label{ss:EnforceLipschitz}

In a second round of experiments, we implemented a model built on  Wasserstein Generative Adversarial Network with Gradient Penalty  \cite{gulrajani2017improved} with a $L^1$-reconstruction loss $\mathcal{L}_{re}$ 
\begin{equation}\label{eq:L1}
    \mathcal{L}_{re}(G)=\frac{1}{N}\sum_{i=1}^N\|S-G(S',A)\|_1.
\end{equation}
in the generator.

\begin{figure}
  \centering
  \begin{tabular}{@{}c@{}}
    \includegraphics[width=0.45\linewidth]{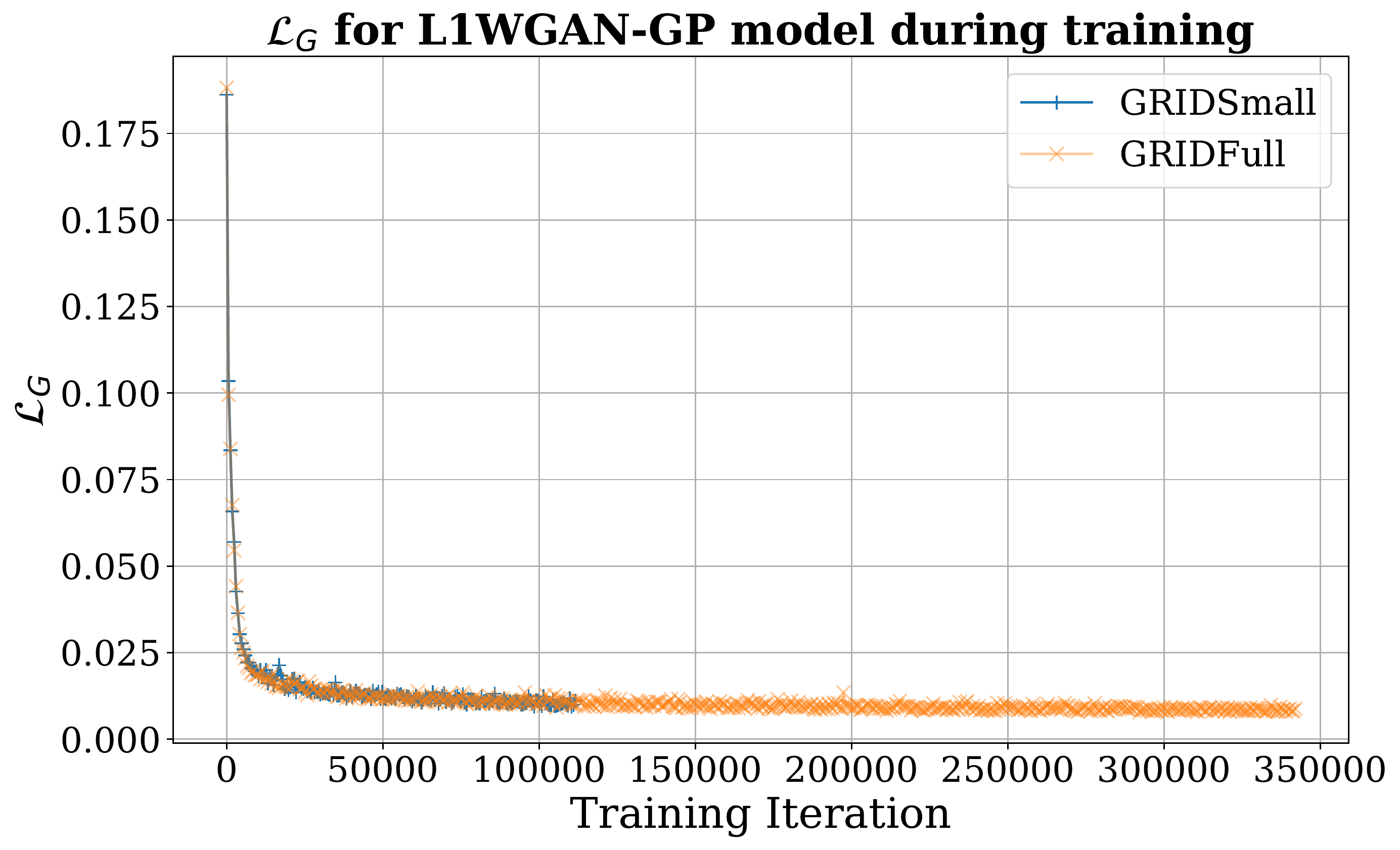} \\[\abovecaptionskip]
    \small (a) Generator loss $\mathcal{L}_G$
  \end{tabular}
  \begin{tabular}{@{}c@{}}
    \includegraphics[width=.45\linewidth]{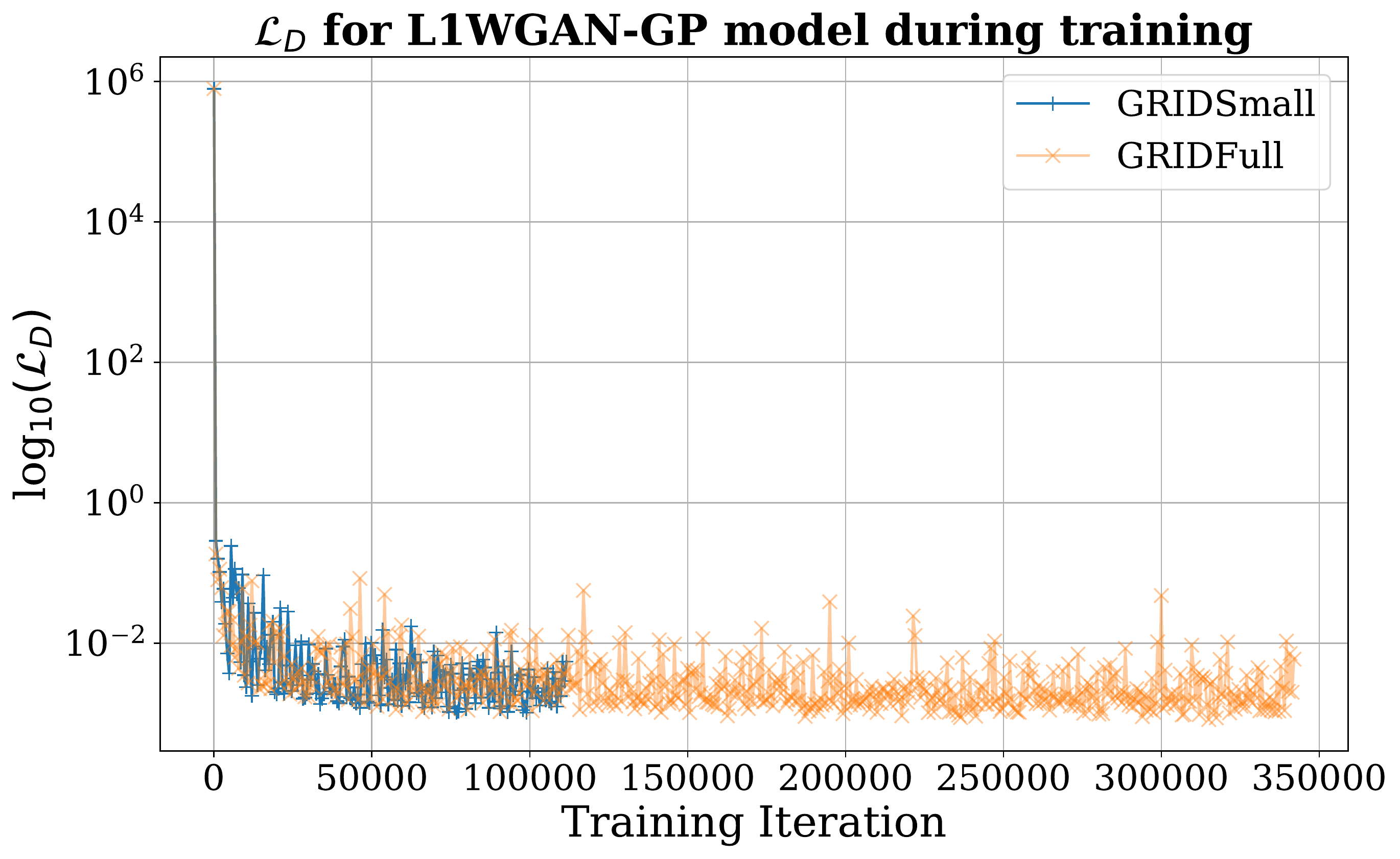} \\[\abovecaptionskip]
    \small (b) Critic loss $\mathcal{L}_D$
  \end{tabular}
      \caption{Losses during training for the WGAN-GP model.}
    \label{fig:L1WGANGPLosses}
\end{figure}

\begin{figure}
    \centering
    \includegraphics[scale=0.3]{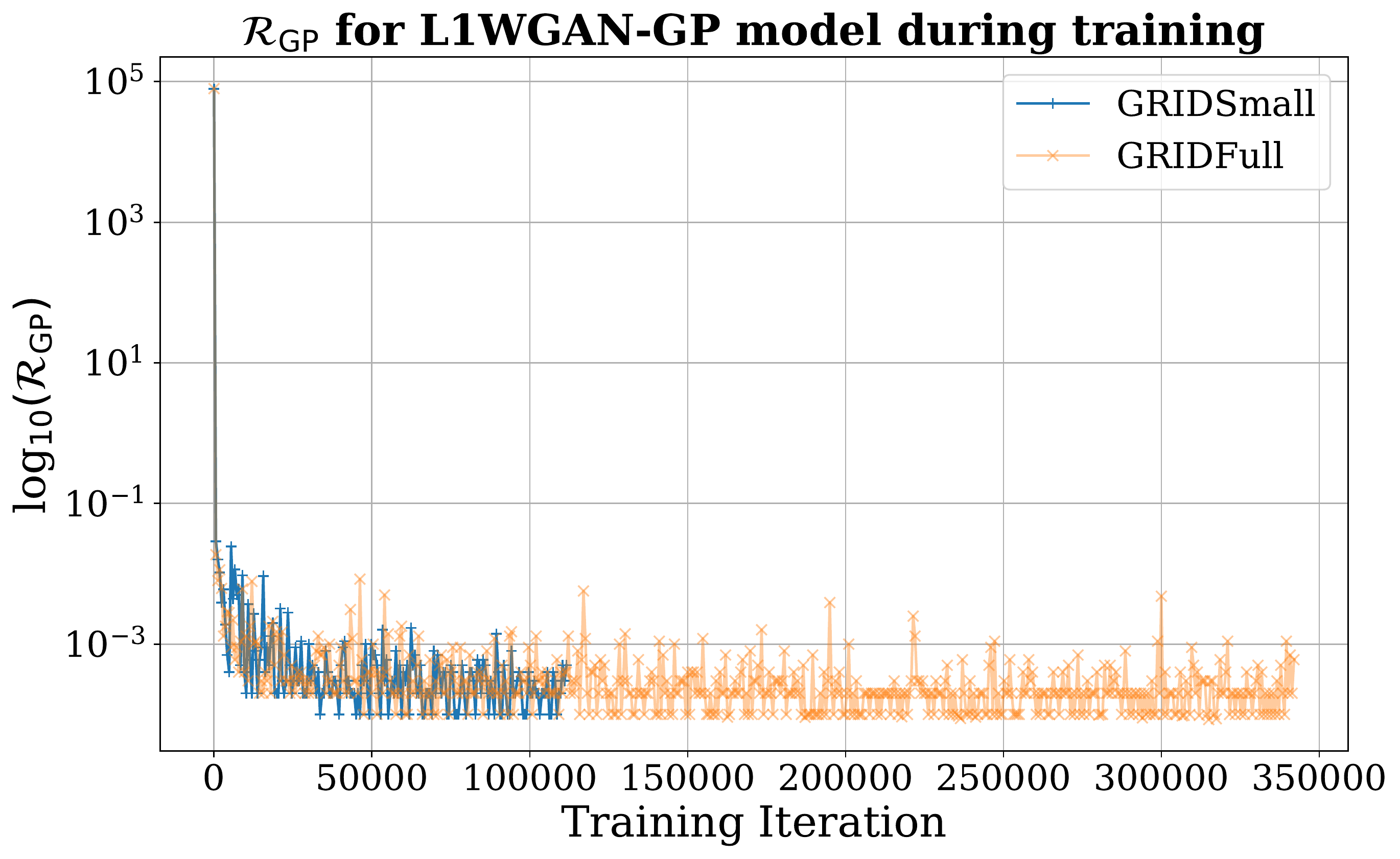}
    \caption{Gradient penalty term $\mathcal{R}_\mathrm{GP}$ for the L1WGAN-GP model.}
    \label{fig:L1WGANGPGradientPenalty}
\end{figure}

\begin{figure}
    \centering
    \includegraphics[scale = 0.17]{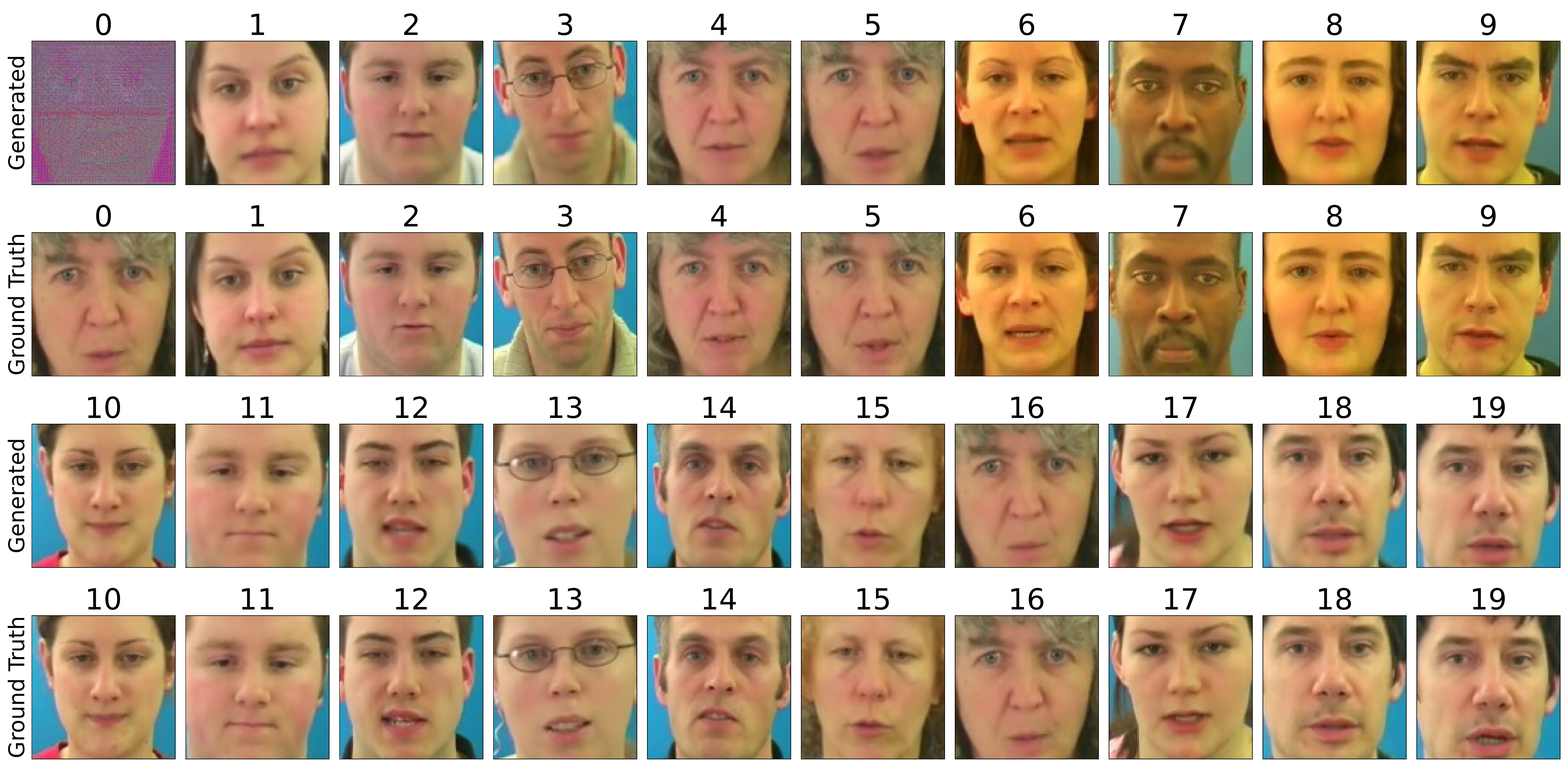}
    \caption{20 random generated faces $\hat{S}$ from the generator, together with their corresponding true faces $S$, from the training of the L1WGAN-GP model using GRIDFull. The epoch for the samples is denoted by the number above each sample.}
    \label{fig:L1WGANGPSamples}
\end{figure}

The gradient penalty is a regularization defined as
\begin{equation} \label{eq:GradientPenalty}
    \mathcal{R}_\mathrm{GP} = \mathbb{E}_{\bm{\hat{x}}\sim\mathbb{P}_{\bm{\hat{x}}}} \bigg [\big (\|\nabla_{\bm{\hat{x}}} D(\bm{\hat{x}})\|_2-1 \big)^2 \bigg]
\end{equation}
where $\bm{\hat{x}}$ is the output of the generator i.e $G(\bm{z})=\bm{\hat{x}}$, $\bm{z}\sim\mathbb{P}_z$. Introducing this term yields a total loss function of
\begin{equation} \label{eq:WGANGPLoss} 
      \mathcal{L}_{\text{WGAN-GP}}(G,D) = \mathbb{E}_{\bm{\hat{x}} \sim \mathbb{P}_g}\bigg[D(\bm{\hat{x}})\bigg]-\mathbb{E}_{\bm{x} \sim \mathbb{P}_r}\bigg[D(\bm{x})\bigg] + \lambda\mathcal{R}_\mathrm{GP}.\end{equation} 
where $\lambda$ is a penalty coefficient and $\bm{\hat{x}}$ the output of the generator i.e $G(\bm{z})=\bm{\hat{x}}$, $\bm{z}\sim\mathbb{P}_z$.
The motivation behind the gradient penalty term in \eqref{eq:WGANGPLoss}, is to penalize gradients with norms differing from 1. 
As the penalty terms for each discriminator input $\bm{\hat{x}}$ are calculated individually, batch normalization can not be used.
The major difference between the WGAN model and the LipGAN model is the training process: for each training step, the WGAN model trained the discriminator every step, and the generator every 5th step, using backpropagation. 
L1WGAN-GP was trained on the GRIDSmall and the GRIDFull dataset for 20 epochs, which resulted in 105000 training iterations and 342400 training iterations respectively. To check convergence, we sampled generator loss $\mathcal{L}_G$ and the discriminator loss $\mathcal{L}_D$ at 600th training iteration, visualized in figure \ref{fig:L1WGANGPLosses}, as well as the gradient penalty term $\mathcal{R}_\mathrm{GP}$, plotted in figure \ref{fig:L1WGANGPGradientPenalty}. 
\noindent \textbf{Sample inspection.}
Samples for the generated faces $\hat{S}$ and their corresponding ground truth part $S$, were saved once per epoch during the training. We see in figure \ref{fig:L1WGANGPSamples} that the samples have a realistic look, but tend to be slightly blurry at times, especially for the early epochs. Upon inference, the model produced distinct faces for each separate frame, and no sign of suspected mode collapse could be observed. 

\noindent

\begin{figure}
  \centering
  \begin{tabular}{@{}c@{}}
    \includegraphics[width=0.45\linewidth]{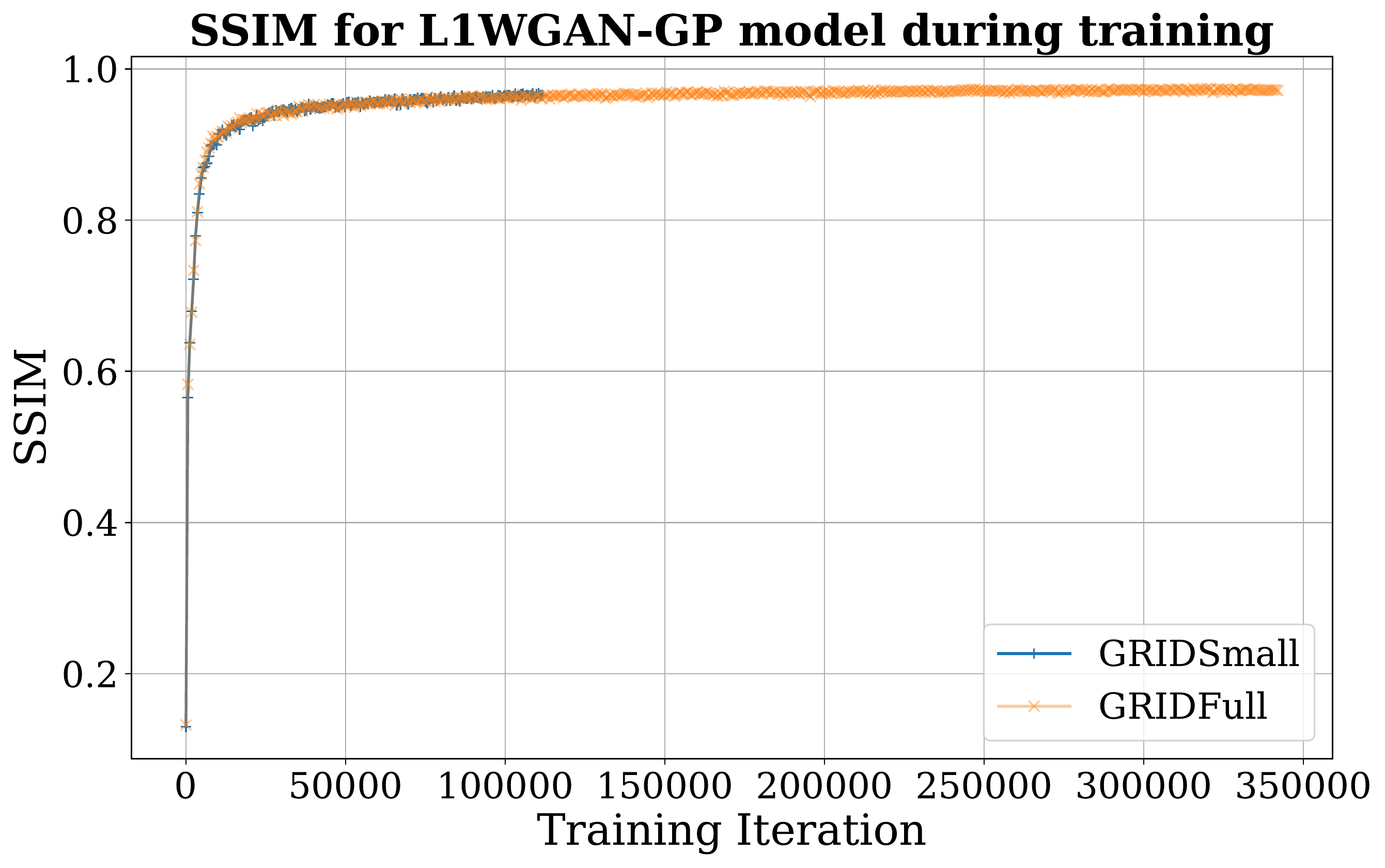} \\[\abovecaptionskip]
    \small (a) SSIM
  \end{tabular}
  \begin{tabular}{@{}c@{}}
    \includegraphics[width=.45\linewidth]{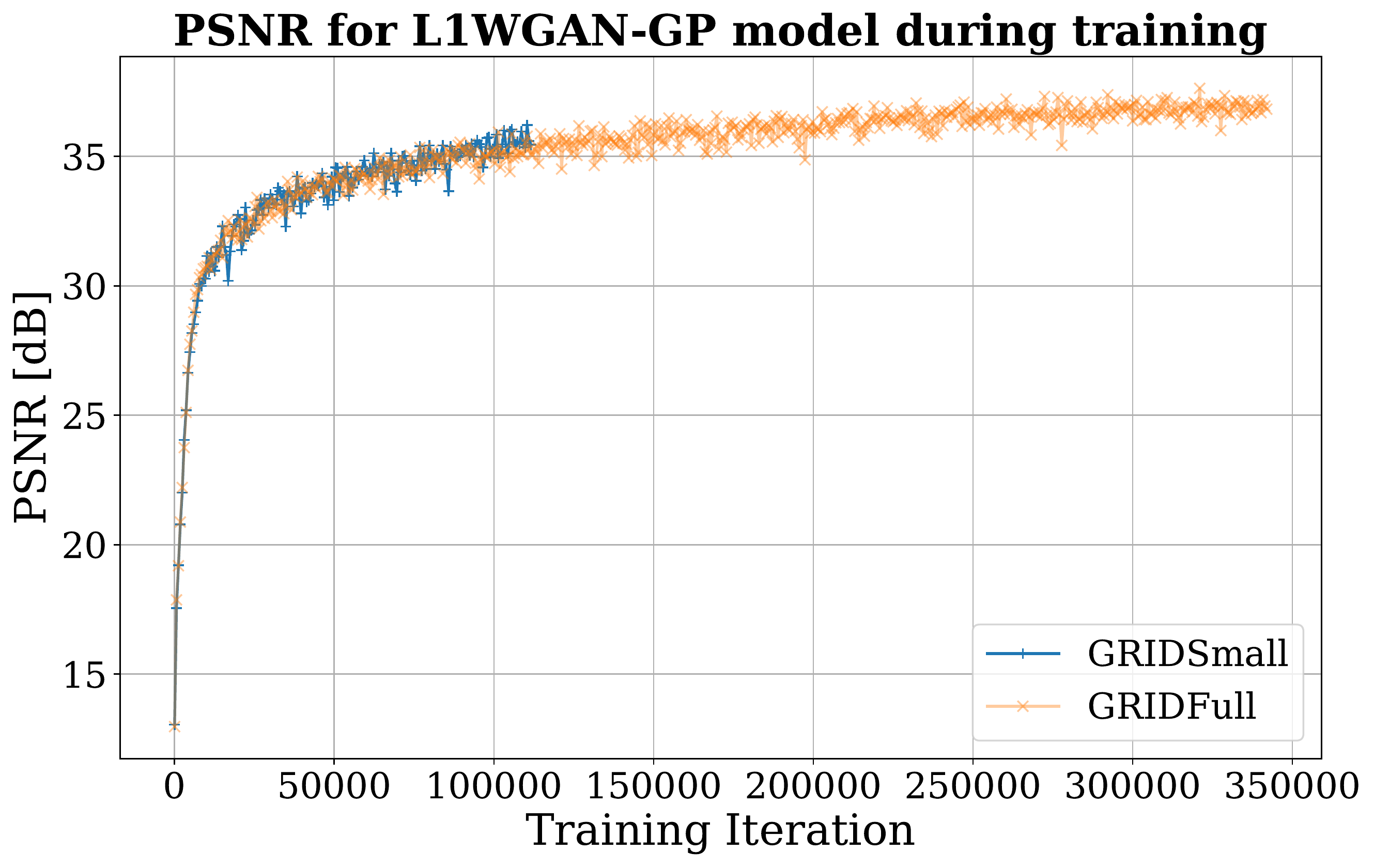} \\[\abovecaptionskip]
    \small (b) PSNR
  \end{tabular}
    \caption{The metrics SSIM and PSNR for the L1WGAN-GP model during training. The metrics have been taken for every 600th training iteration.}
    \label{fig:L1WGANGPTrainingMetrics}
\end{figure}

\noindent \textbf{Quality metrics.} As visualized in figure \ref{fig:L1WGANGPTrainingMetrics}, the SSIM goes from approximately 0.13 for both models to 0.97 and 0.98 for GRIDSmall and GRIDFUll respectively. Further, the PSNR ranges from around 13 dB for both datasets, to around 36 dB for GRIDSmall, and 37 dB for GRIDFull.

\section{Results}
In this section we summarize the results of our experiments.
The motivation for our work was the animation of an image, showing a portrait-snipped of a single person on ``good'' background, into a short video message with prescribed audio. For this task, the GRID dataset seemed to be most adapted, given (a) its controlled setting and (b) its annotated audio transcription, making it most convenient for a later implementation of a text-to-speech feature.

\subsection{Dataset impact on LipGAN}
\begin{figure}
  \centering
  \begin{tabular}{@{}c@{}}
    \includegraphics[width=0.45\linewidth]{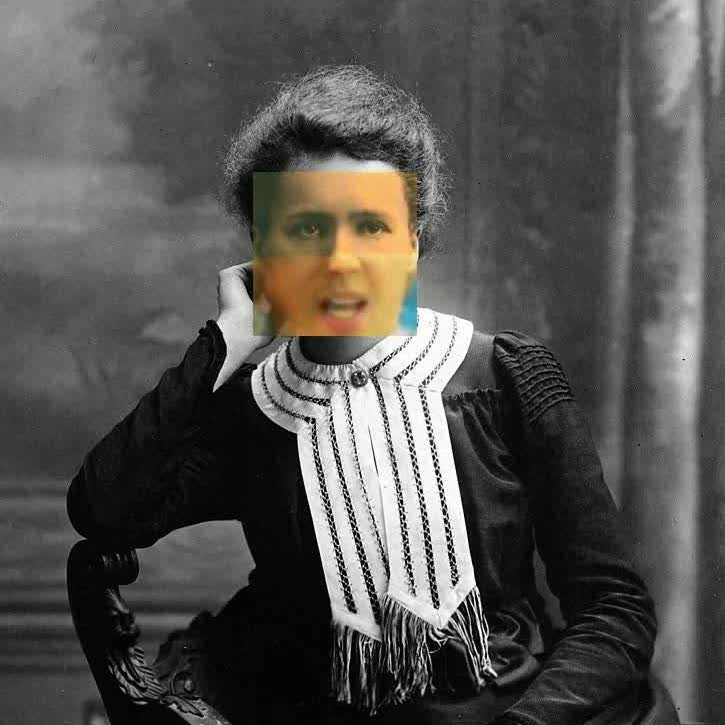} \\[\abovecaptionskip]
    \small (a) LipGAN trained on GRID.
  \end{tabular}
  \begin{tabular}{@{}c@{}}
    \includegraphics[width=.45\linewidth]{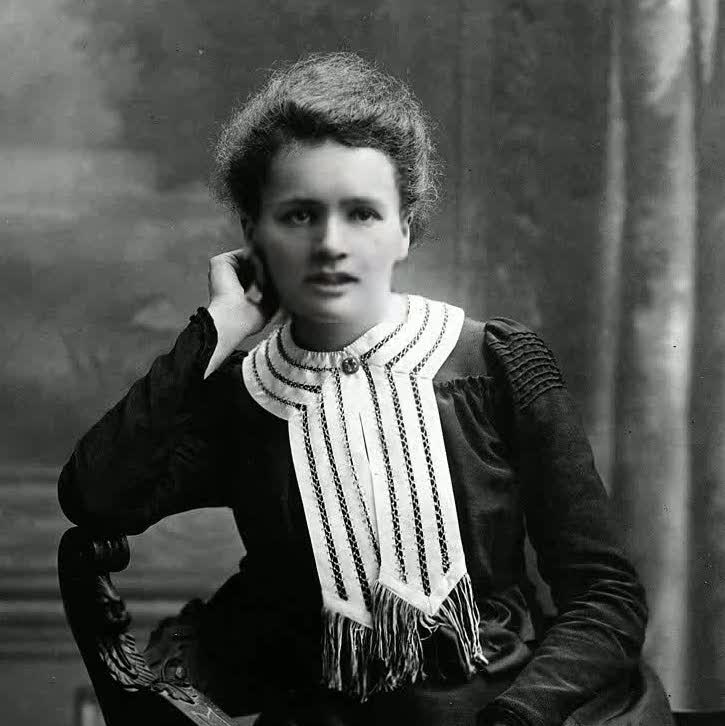} \\[\abovecaptionskip]
    \small (b) LipGAN trained on LRS2.
  \end{tabular}
    \caption{Inference of LipGAN trained on two different datasets. The LipGAN model trained with GRID did not manage to adapt to the color scheme of the target image.  Image source: Wikipedia.}
    \label{fig:curie}
\end{figure}

A first surprising outcome is the poor generalization of our re-implemented Pytorch LipGAN to grayscale images. While 
the original Keras-LipGAN model trained on LRS2 gave  satisfactory visual results after inference,  the Pytorch-LipGAN model trained on GRID did not manage to adapt to the new color scheme, as showcased in figure \ref{fig:curie}. 
One might also speculate whether the mouth is slighly misplaced and slighly larger than it should. 
\noindent 
We conclude that the outcome of inference is sensitively depending on the properties of the used target data. This problem is typical for GAN algorithms, and also discussed by \cite{vougioukas2020realistic}, whose model performance is also limited to quite controlled settings of well-aligned frontal faces, such as in GRID. 

Further studies on a data-augmented GRID dataset with additional grayscale images should be performed to investigate the issue, and see if it can be remedied for LipGAN trained on GRID.

\subsection{FID, SSIM and PSNR scores}

We compare LipGAN and L1WGAN-GP  in terms of three quantitative metrics, FID, SSIM and PSNR, evaluated on unseen test data, in the form of the GRIDTest dataset.  
All scores used the $44589$ points of reference data in TestGRID; SSIM and PSNR are visualized as boxplots with outliers omitted for better visibility 
%
The results of the FID score are presented in table \ref{tab:FIDResult}. Here, the L1WGAN-GP model outperforms our re-implementation of LipGAN, signifying that, for L1WGAN-GP, the generated data distribution is closer to the reference data distribution. 

\begin{table}
\centering
\caption{FID-score on the models trained using GRIDFull.}
\begin{tabular}{c|c}
\hline \hline
\textbf{Model} & \textbf{FID-Score $\downarrow$} \\ \hline
LipGAN         & 15.11                           \\
L1WGAN-GP      & \textbf{14.49} \\ \hline \hline
\end{tabular}
\label{tab:FIDResult}
\end{table}

In terms of SSIM, both LipGAN and L1WGAN-GP reach values close to the maximum possible value of $1.0$, with  LipGAN performing slightly better than L1WGAN-GP in terms of both median and mean value. 
Table \ref{tab:SSIMResult} summarizes the numeric properties of the acquired SSIM scores. 
\begin{table}[H]
\centering
\caption{SSIM summary statistics for the models trained on GRIDFull.}
\begin{tabular}{c|cccc}
\hline \hline
\textbf{Model} & \textbf{Mean $\uparrow$}   & \textbf{Median $\uparrow$} & \textbf{Max $\uparrow$}    & \textbf{Min $\uparrow$}    \\ \hline
LipGAN         & \textbf{0.9348} & \textbf{0.9439} & \textbf{0.9796} & \textbf{0.7542} \\
L1WGAN-GP      & 0.9296          & 0.9380          & 0.9754          & 0.7052          \\ \hline \hline
\end{tabular}
\label{tab:SSIMResult}
\end{table}
\begin{figure}
    \centering
    \includegraphics[scale=0.5]{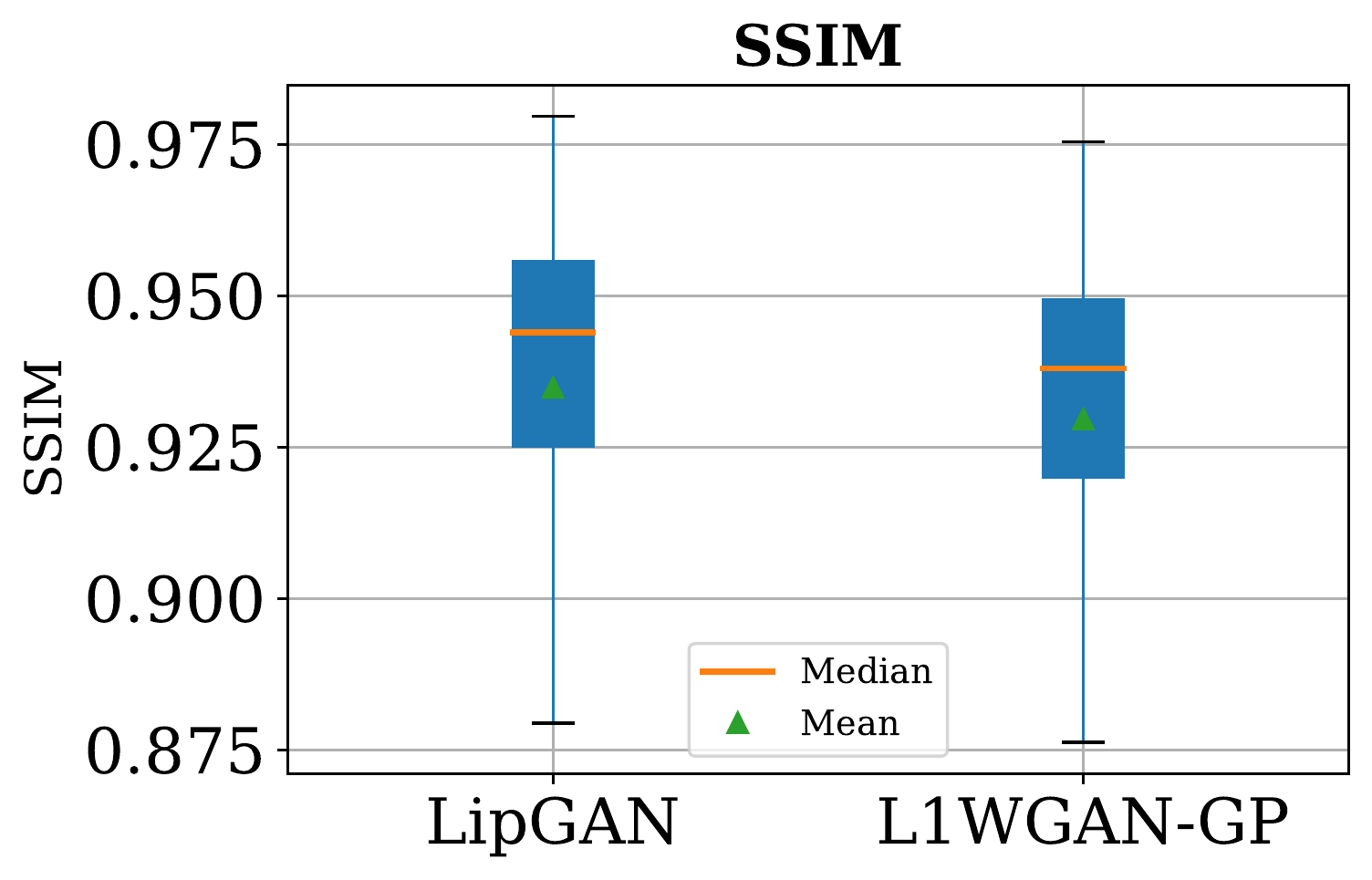}
    \caption{SSIM for the models trained on GRIDFull. Outliers have been omitted from the box plot.}
    \label{fig:compare_ssim}
\end{figure}
\noindent
%
%
%
%
\begin{figure}
    \centering
    \includegraphics[scale=0.5]{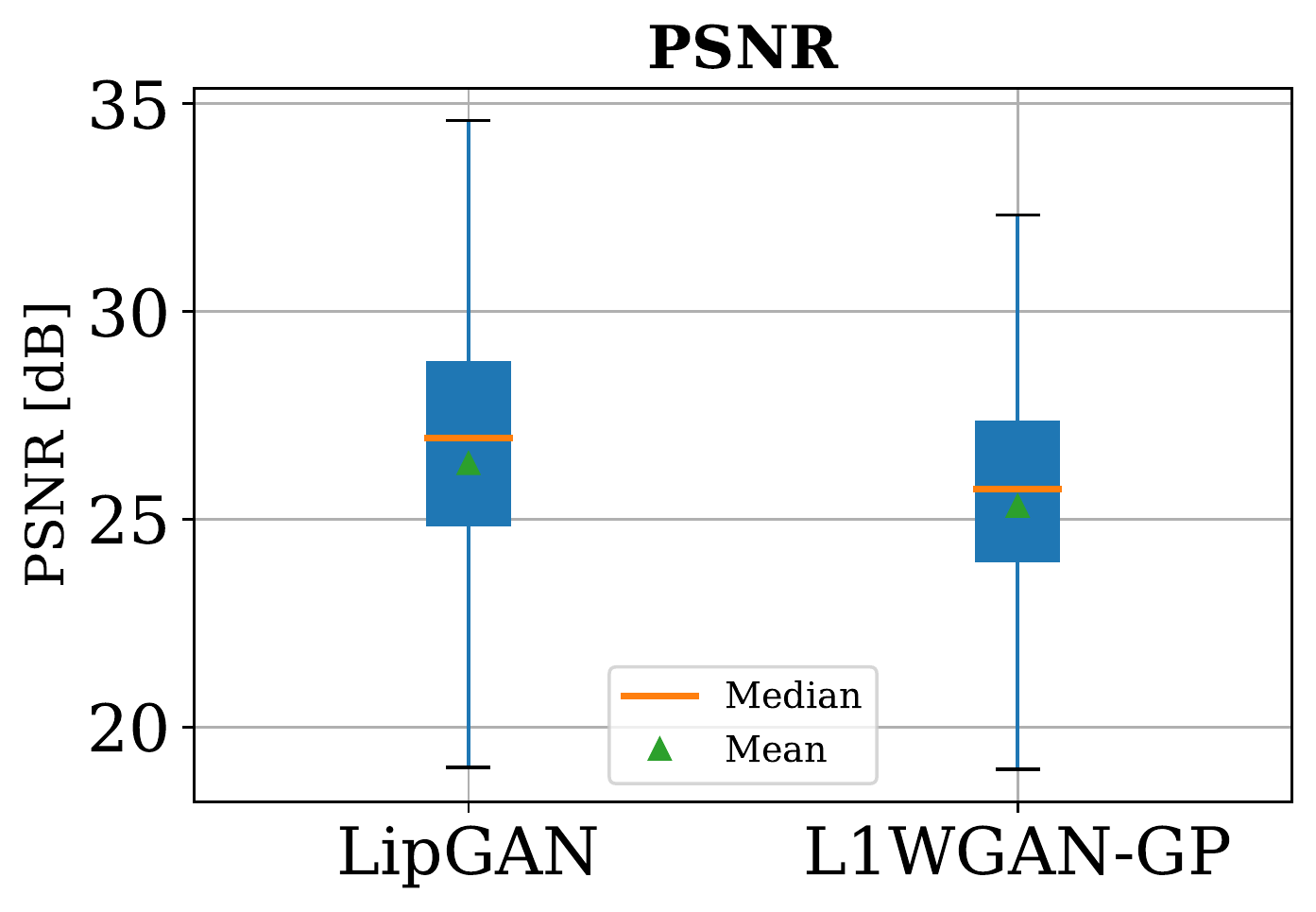}
    \caption{PSNR for the models trained using GRIDFull. Outliers have been omitted from the box plot.}
    \label{fig:compare_psnr}
\end{figure}
\noindent
For PSNR, we see a very different phenomenon: first, the spread was quite large, ranging around $15$ dB, and LipGAN outperformed L1WGAN-GP in terms of median and mean.
\begin{table}
\centering
\caption{PSNR summary statistics for the models trained using GRIDFull.}
\begin{tabular}{c|cccc}
\hline \hline
\textbf{Model} & \textbf{Mean {[}dB{]} $\uparrow$} & \textbf{Median {[}dB{]} $\uparrow$} & \textbf{Max {[}dB{]} $\uparrow$} & \textbf{Min {[}dB{]} $\uparrow$} \\ \hline 
LipGAN         & \textbf{26.34}         & \textbf{26.96}           & \textbf{35.35}        & \textbf{13.67}        \\
L1WGAN-GP      & 25.32                  & 25.72                    & 34.84                 & 12.81                 \\ \hline \hline
\end{tabular}
\label{tab:PSNRResult}
\end{table}

\subsection{Qualitative comparison}
\begin{figure}
  \centering
  \begin{tabular}{@{}c@{}}
    \includegraphics[width=0.45\linewidth]{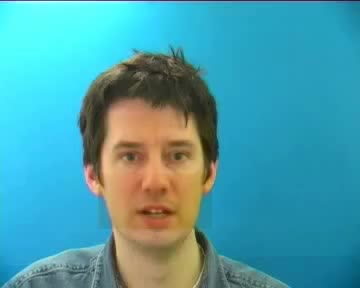} \\[\abovecaptionskip]
    \small (a) Inference sample from LipGAN.
  \end{tabular}
  \begin{tabular}{@{}c@{}}
    \includegraphics[width=.45\linewidth]{L1WGAN-GP_box.jpg} \\[\abovecaptionskip]
    \small (b) Inference sample from L1WGAN-GP.
  \end{tabular}
  \caption{Example of a visual box surrounding the face during inference. This phenomenon occurred for both models.}
    \label{fig:box}
\end{figure}

\begin{figure}
    \centering
    \includegraphics[scale=0.2, clip=true, trim=0in 5.5in 0in 5.5in]{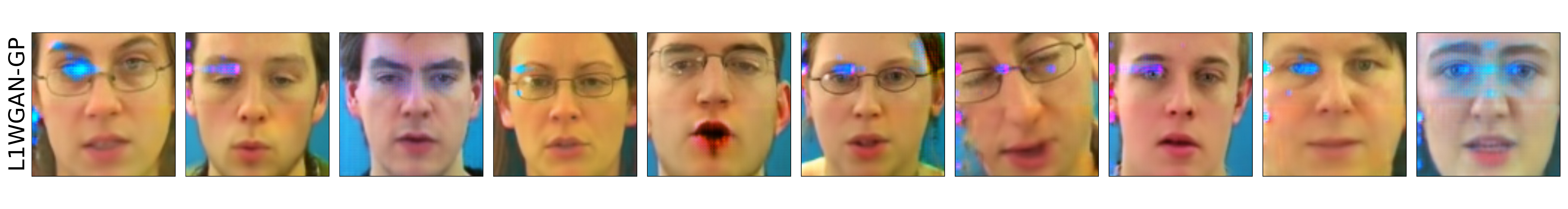}
    \caption{Example of visible artifacts from images produced by the L1WGAN-GP model with GRIDTrain as input.}
    \label{fig:collageArtefacts}
\end{figure}

We further examine some qualitative aspects of the two models by looking at the generated data produced by using GRIDTrain as input. 
As a first remark, it should be highlighted that both models solved the task of lip-synchronization adequately good in a subjective manner.
We note upon inspection of data produced during inference, of both models, a certain discrepancy between the generated face and the background. This can be noticed as a visible box surrounding the face. This phenomenon was noticed equally much for both models and is displayed in figure \ref{fig:box}.
Additionally, some of the images produced by
L1WGAN-GP  had visual artifacts, see examples in figure \ref{fig:collageArtefacts}. This problem was not experienced in data produced by LipGAN.
\noindent
 The artefacts in L1WGAN-GP  mostly occurr around the eyes of the target face and appear visually in a variety of ways, as discolored pixels,  in many cases matching the surrounding background, see figure \ref{fig:collageArtefacts}. The artifacts most likely originate from the fact that the L1WGAN-GP model fails to differentiate the background from certain areas of the face.
It is difficult to determine why these artefacts appear when trained with GRID, which is a relatively controlled dataset. To some extend, artefacts are unavoidable when facial expressions need to be generated, as these always imply deviations from the ground truth. It could be that more training hours are needed, as the generator in L1WGAN-GP only begins updated every fifth iteration. However, this is ruled out by the quick convergence of losses and similar metric performance compared to LipGAN.

\section{Conclusion and outlook}

To summarize, the quantitative metrics used were not very conclusive when comparing  LipGAN and L1WGAN-GP. In all three cases, the results were numerically close to each other,  while sample inspection did reveal flaws in  L1WGAN-GP:  a large number of artifacts were noticed in the images produced by the L1WGAN-GP model. 

Considering the artifacts produced by L1WGAN-GP, one would have wished for a larger discrepancy in the quantitative metric scores. 
We guess that this did not happen as the metrics compare entire pictures, thus a small artifact would not render a large difference in the metrics. In contrast to other applications, for the lip-synchronization task, which focusses on a quite small region in the image,  even small artefacts can ruin the human-perceived image quality.  

Furthermore, the focus of quantitative metrics on image quality and congruence with the ground truth makes them unsatisfactory for animation tasks, as it was outlined already by several authors. 

While there is already no consensus about adequat quantitative metrics for GANs that output images, it seems to be even more challenging to determine a proper quantitative scoring system to measure the quality of video output. 

In researchers's attempts to make human-eye qualitative assessment more 'standardized', large-scale versions the former were implemented, such as ``online Turing tests''\cite{vougioukas2020realistic} or mean-opinion-score. There is, however, since years warnings \cite{Viswanathan2005,streijl2016mean} that it is not clear what MOS actually measures, as no dimensions of output quality and no standardization of tester scores is ensured.

We conclude that there is currently no appropriate alternative to human inspection available to quantitatively measure the quality of lip-synchronization, not because there would not be research on multidimensional quality measurement alternatives \cite{streijl2016mean}, but simply because a single number seems to be given preference to a more sophisticated analysis.

\section{Acknowledgements}
PN and JL thank Michael Truong, Simon Akesson and Pieter Buteneers for helpful discussions and Sinch AB Malm\"o lab for providing computational resources.

\bibliographystyle{plain}

\begin{thebibliography}{10}

\bibitem{Afouras_2019}
Triantafyllos Afouras, Joon~Son Chung, Andrew Senior, Oriol Vinyals, and Andrew
  Zisserman.
\newblock Deep audio-visual speech recognition.
\newblock {\em IEEE Transactions on Pattern Analysis and Machine Intelligence},
  page 1–1, 2019.

\bibitem{chen2018}
Lele Chen, Zhiheng Li, Ross~K Maddox, Zhiyao Duan, and Chenliang Xu.
\newblock Lip movements generation at a glance.
\newblock In {\em Proceedings of the European Conference on Computer Vision
  (ECCV)}, pages 520--535, 2018.

\bibitem{chong2020effectively}
Min~Jin Chong and David Forsyth.
\newblock Effectively unbiased fid and inception score and where to find them,
  2020.

\bibitem{chung2017said}
Joon~Son Chung, Amir Jamaludin, and Andrew Zisserman.
\newblock You said that?
\newblock {\em CoRR}, abs/1705.02966, 2017.

\bibitem{cooke2006audio}
Martin Cooke, Jon Barker, Stuart Cunningham, and Xu~Shao.
\newblock An audio-visual corpus for speech perception and automatic speech
  recognition.
\newblock {\em The Journal of the Acoustical Society of America},
  120(5):2421--2424, 2006.

\bibitem{fan2015photo}
Bo~Fan, Lijuan Wang, Frank~K Soong, and Lei Xie.
\newblock Photo-real talking head with deep bidirectional lstm.
\newblock In {\em 2015 IEEE International Conference on Acoustics, Speech and
  Signal Processing (ICASSP)}, pages 4884--4888. IEEE, 2015.

\bibitem{gulrajani2017improved}
Ishaan Gulrajani, Faruk Ahmed, Martin Arjovsky, Vincent Dumoulin, and Aaron~C
  Courville.
\newblock Improved training of wasserstein gans.
\newblock {\em Advances in neural information processing systems}, 30, 2017.

\bibitem{kumar2017obamanet}
Rithesh Kumar, Jose Sotelo, Kundan Kumar, Alexandre de~Brebisson, and Yoshua
  Bengio.
\newblock Obamanet: Photo-realistic lip-sync from text, 2017.

\bibitem{kurach2019largescale}
Karol Kurach, Mario Lucic, Xiaohua Zhai, Marcin Michalski, and Sylvain Gelly.
\newblock A large-scale study on regularization and normalization in gans,
  2019.

\bibitem{liu2017video}
Ziwei Liu, Raymond~A Yeh, Xiaoou Tang, Yiming Liu, and Aseem Agarwala.
\newblock Video frame synthesis using deep voxel flow.
\newblock In {\em Proceedings of the IEEE international conference on computer
  vision}, pages 4463--4471, 2017.

\bibitem{LipGANArticle}
K~R Prajwal, Rudrabha Mukhopadhyay, Philip Jerin, Jha Abhishek, Vinay
  Namboodiri, and C~V Jawahar.
\newblock Towards automatic face-to-face translation.
\newblock In {\em Proceedings of the 27th ACM International Conference on
  Multimedia}, MM '19, pages 1428--1436, New York, NY, USA, 2019. ACM.

\bibitem{simons1990generation}
A.D. Simons and SJ. Cox.
\newblock Generation of mouthshape for a synthetic talking head.
\newblock {\em Proc. of the Institute of Acoustics}, 1990.

\bibitem{streijl2016mean}
Robert~C Streijl, Stefan Winkler, and David~S Hands.
\newblock Mean opinion score (mos) revisited: methods and applications,
  limitations and alternatives.
\newblock {\em Multimedia Systems}, 22(2):213--227, 2016.

\bibitem{suwajanakorn2017synthesizing}
Supasorn Suwajanakorn, Steven~M Seitz, and Ira Kemelmacher-Shlizerman.
\newblock Synthesizing obama: learning lip sync from audio.
\newblock {\em ACM Transactions on Graphics (ToG)}, 36(4):1--13, 2017.

\bibitem{thies2016face2face}
Justus Thies, Michael Zollhofer, Marc Stamminger, Christian Theobalt, and
  Matthias Nie{\ss}ner.
\newblock Face2face: Real-time face capture and reenactment of rgb videos.
\newblock In {\em Proceedings of the IEEE conference on computer vision and
  pattern recognition}, pages 2387--2395, 2016.

\bibitem{Viswanathan2005}
Mahesh Viswanathan and Madhubalan Viswanathan.
\newblock Measuring speech quality for text-to-speech systems: development and
  assessment of a modified mean opinion score (mos) scale.
\newblock {\em Computer Speech \& Language}, 19(1):55--83, 2005.

\bibitem{vougioukas2020realistic}
Konstantinos Vougioukas, Stavros Petridis, and Maja Pantic.
\newblock Realistic speech-driven facial animation with gans.
\newblock {\em International Journal of Computer Vision}, 128(5):1398--1413,
  2020.

\bibitem{wang2010synthesizing}
Lijuan Wang, Xiaojun Qian, Wei Han, and Frank~K Soong.
\newblock Synthesizing photo-real talking head via trajectory-guided sample
  selection.
\newblock In {\em Eleventh Annual Conference of the International Speech
  Communication Association}, 2010.

\bibitem{wiles2018x2face}
Olivia Wiles, A~Koepke, and Andrew Zisserman.
\newblock X2face: A network for controlling face generation using images,
  audio, and pose codes.
\newblock In {\em Proceedings of the European conference on computer vision
  (ECCV)}, pages 670--686, 2018.

\bibitem{xie2007realistic}
Lei Xie and Zhi-Qiang Liu.
\newblock Realistic mouth-synching for speech-driven talking face using
  articulatory modelling.
\newblock {\em IEEE Transactions on Multimedia}, 9(3):500--510, 2007.

\bibitem{xu2018evaluation}
Qiantong Xu, Gao Huang, Yang Yuan, Chuan Guo, Yu~Sun, Felix Wu, and Kilian
  Weinberger.
\newblock An empirical study on evaluation metrics of generative adversarial
  networks.
\newblock {\em arXiv preprint arXiv:1806.07755}, 2018.

\bibitem{yao2021iterative}
Xinwei Yao, Ohad Fried, Kayvon Fatahalian, and Maneesh Agrawala.
\newblock Iterative text-based editing of talking-heads using neural
  retargeting.
\newblock {\em ACM Transactions on Graphics (TOG)}, 40(3):1--14, 2021.

\bibitem{zhou2019talking}
Hang Zhou, Yu~Liu, Ziwei Liu, Ping Luo, and Xiaogang Wang.
\newblock Talking face generation by adversarially disentangled audio-visual
  representation.
\newblock In {\em Proceedings of the AAAI conference on artificial
  intelligence}, volume~33, pages 9299--9306, 2019.

\end{thebibliography}

\end{document}